%% file: acl_latex.tex
\pdfoutput=1

\documentclass[11pt]{article}

\usepackage{acl}

\usepackage{times}
\usepackage{latexsym}

\usepackage[T1]{fontenc}

\usepackage[utf8]{inputenc}

\usepackage{microtype}
\usepackage[compact]{titlesec}
\usepackage{bm}
\usepackage{todonotes}
\usepackage{amsmath,amssymb,amsfonts}
\usepackage{graphicx,booktabs,adjustbox}

\newcommand{\specialcell}[2][c]{%
  \begin{tabular}[#1]{@{}c@{}}#2\end{tabular}}
\newcommand{\sub}[1]{\textsubscript{#1}}
\newcommand{\sups}[1]{\textsuperscript{#1}}

\newcommand{\sigbertlow}{\sups{$\circ$}}
\newcommand{\sigberthigh}{\sups{$\bullet$}}
\newcommand{\sigbertsoclow}{\sups{$\vartriangle$}}
\newcommand{\sigbertsochigh}{\sups{$\blacktriangle$}}

\usepackage{xspace}
\newcommand{\placeholder}{BERT+EAR\xspace}
\newcommand{\reg}{EAR\xspace}

\newcommand{\repository}{\url{https://github.com/g8a9/ear}\xspace}

\title{Entropy-based Attention Regularization\\Frees Unintended Bias Mitigation from Lists}

 \author{Giuseppe Attanasio$^{1,2}$, Debora Nozza$^{1}$, Dirk Hovy$^{1}$, Elena Baralis$^{2}$ \\ \\
 $^1$Bocconi University, Milan, Italy \\
 $^2$Politecnico di Torino, Turin, Italy \\
 { \tt \{giuseppe.attanasio3,debora.nozza,dirk.hovy\}@unibocconi.it}, \\
 { \tt elena.baralis@polito.it}}

\begin{document}
\maketitle
\begin{abstract}
\emph{Warning: This paper contains examples of language that some people may find offensive.}

Natural Language Processing (NLP) models risk overfitting to specific terms in the training data, thereby reducing their performance, fairness, and generalizability. E.g., neural hate speech detection models are strongly influenced by identity terms like \textit{gay}, or \textit{women}, resulting in false positives, severe unintended bias, and lower performance.
Most mitigation techniques use lists of identity terms or samples from the target domain during training. However, this approach requires a-priori knowledge and introduces further bias if important terms are neglected.
Instead, we propose a knowledge-free Entropy-based Attention Regularization (EAR) to discourage overfitting to training-specific terms. An additional objective function penalizes tokens with low self-attention entropy.
We fine-tune BERT via EAR: the resulting model matches or exceeds state-of-the-art performance for hate speech classification and bias metrics on three benchmark corpora in English and Italian.
EAR also reveals overfitting terms, i.e., terms most likely to induce bias, to help identify their effect on the model, task, and predictions.
\end{abstract}

\section{Introduction}

\begin{figure}[!tp]
    \centering
    \includegraphics[width= .80 \linewidth]{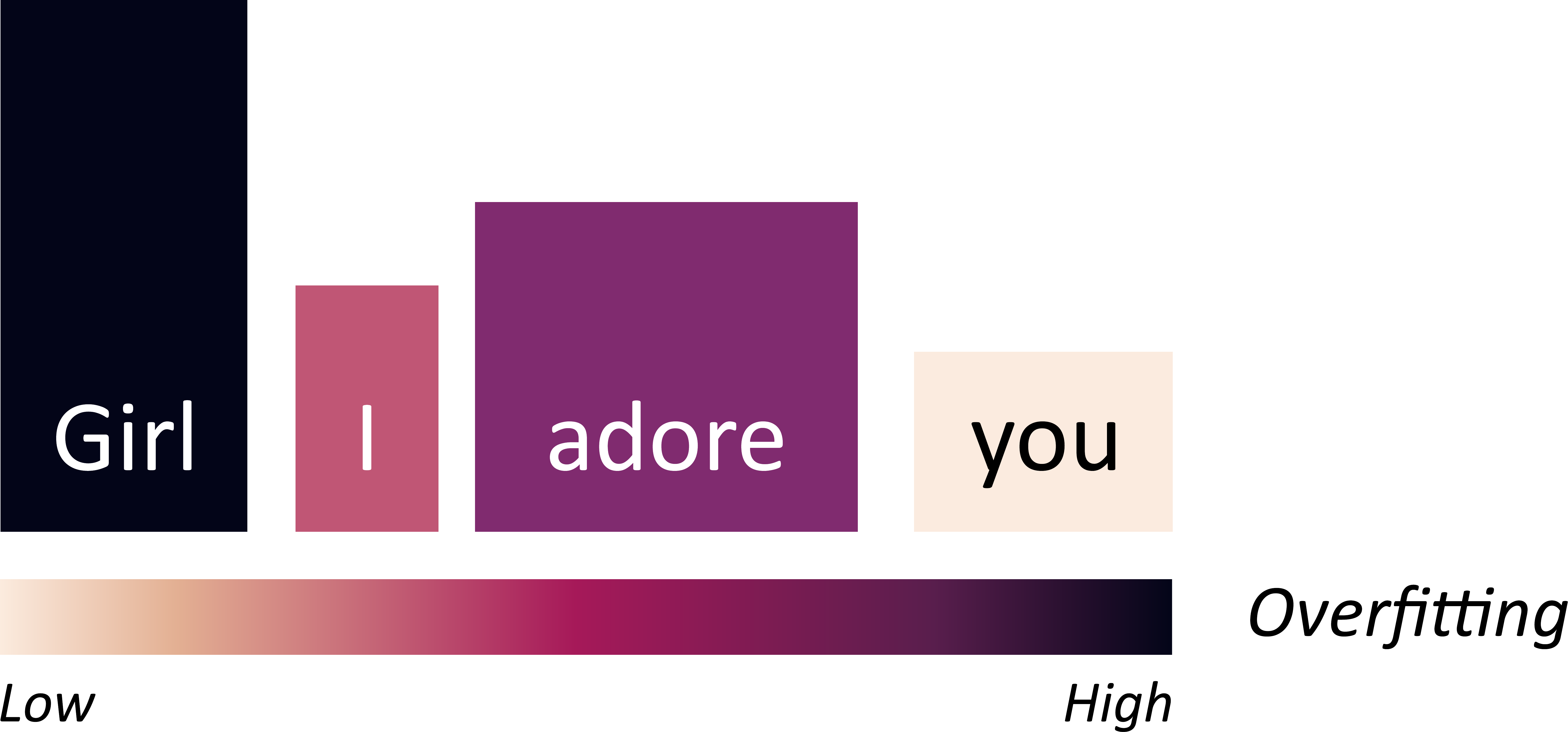}
    \caption{False positive from BERT as a hate speech detector. The darker and taller the bar, the higher the overfitting on the term.}
    \label{fig:running_example}
\end{figure}

Online hate speech is growing at a rapid pace, with effects that can result in dangerous criminal acts offline. Due to its verbal nature, various Natural Language Processing approaches have been proposed \cite[inter alia]{qian-etal-2018-hierarchical,indurthi-etal-2019-fermi,attanasio_politeam_2020,kennedy-etal-2020-contextualizing,vidgen-etal-2021-learning}.
Recently, detection performance has significantly improved with the use of large pre-trained language models based on Transformers \cite{NIPS2017_3f5ee243}, such as Bidirectional Encoder Representations from Transformers (BERT)~\cite{devlin-etal-2019-bert}.
However, several works have shown that by fine-tuning neural language models on hate speech detection, the classifiers obtained contain severe \textit{unintended bias} \cite{dixon2018}, i.e. they perform better or worse when texts mention specific \textit{identity terms} (such as \textit{gay}, \textit{Muslim}, or \textit{woman}). As a result, a sentence like ``As a Muslim woman, I agree'' would be wrongly classified as hate speech, purely due to the presence of two identity terms, i.e., terms referring to specific groups based on their socio-demographic features. 
One cause of false positives is selection bias in the keyword-driven collection of corpora \cite{ousidhoum-etal-2020-comparative}.
Figure~\ref{fig:running_example} shows a false positive example for a fine-tuned BERT model on hate speech detection. Ideally, the model should rely on the words \textit{adore} and \textit{you}. Instead, BERT overfitted to the word \textit{Girl} and associated it with a hateful context. This unwanted effect demonstrates the issues of lexical overfitting, and how they cause uninteded bias on identity terms.

Various methods have been proposed to mitigate and measure (unintended) bias \cite{elazar-goldberg-2018-adversarial,park-etal-2018-reducing,dixon2018,nozza2019unintended,kennedy-etal-2020-contextualizing,vaidya2020empirical}.
However, all those methods rely on the availability of a set of \textit{identity terms}. This is a severe limitation, which hinders the generalizability and applicability of hate detection models to real-world contexts. For example, a model designed to reduce the unintended bias on gender-related terms (such as \textit{woman}, \textit{wife}) will not address unintended bias for religious affiliation. So practitioners must decide a-priori \textit{``which vulnerable groups are present in our data?''}

We propose an Entropy-based Attention Regularization (\reg) that forces the model to build token representations by attending to a wider context, i.e., consider a larger number of tokens from the rest of the sentence. We measure the attended context as the entropy of the self-attention weight distribution over the input sequence. We use \reg as a regularization term in the loss computation to maximize each token's entropy.
We apply \reg to BERT. The resulting model (\placeholder) significantly improves performance on unintended bias mitigation in English and Italian. In addition, it requires no a-priori knowledge (e.g., sets of identity terms), making it fairer and more general. The contextualized representations \reg induces avoid basing the classification on individual terms and, ultimately, mitigate lexical overfitting and intrinsic bias from pre-trained weights.

As a training by-product, \reg lets us extract the overfitting terms, i.e., terms accounting for narrower context that most likely induce unintended bias.
These terms can highlight possible weaknesses in the model: from the over-sensitivity of pre-trained weights to specific words~\cite{sheng-etal-2019-woman,nangia-etal-2020-crows,vig2020investigating}, to over-specialization of training corpora on the keywords used for collecting data \cite{ousidhoum-etal-2020-comparative}.

Note that while we show results on BERT, \reg is applicable to any attention-based architecture.




\paragraph{Contributions.}

\reg is a novel entropy-based attention regularization method to mitigate unintended bias by reducing lexical overfitting. 
It is applied to all terms, so it \textit{does not need a-priori domain knowledge} (e.g, predefined term lists).
Independent of domain-specific information, \reg \textit{generalizes better to different languages and contexts} compared to similar approaches. Attention entropy is used to extract a list of the most likely \textit{biased terms}. 
\reg code is available at \repository.

\section{Entropy-based Attention Regularization}
Attention was originally designed for aligning target and source sequences in machine translation \cite{DBLP:journals/corr/Graves13, DBLP:journals/corr/BahdanauCB14}. 
However, in the Transformer architecture \cite{NIPS2017_3f5ee243}, it has become a means to account for lexical influence and long-range dependencies. 
It also provides useful information about the importance of a term for the output \cite{wiegreffe-pinter-2019-attention, Brunner2020On, sun-marasovic-2021-effective}.
Here, we use the notion of attention entropy, and \reg's use of it in BERT. Note, though, that \reg can be used with \textit{any} attention-based architecture.

\paragraph{Self-attention in Transformers.}
The Transformer model consists of two connected units, an encoder and a decoder, designed for sequence-to-sequence tasks. 

A transformer encoder applies scaled-dot product self-attention over the input tokens to compute $N$ independent attention heads.\footnote{In the following, we use \textit{token} and \textit{embedding} interchangeably. We represent vectors with lowercase bold letters.}
Let $E = [{\bm{e_0},...,\bm{e_{d_s}}}]$ be the sequence of input embeddings, with $\bm{e_i} \in \mathbb{R}^{d_m}$.
For the \textit{h}-th attention head and \textit{i}-th position, each embedding $\bm{e_i}$ is projected into a query $\bm{q_{h,i}}$, a key $\bm{k_{h,i}}$ and value $\bm{v_{h,i}}$. So each token expresses an attention distribution over all input embeddings as
\begin{equation}
    a_{h,i} = \operatorname{softmax}\left(\frac{\bm{q_{h,i}}^{T} K_h}{\sqrt{d_{k}}}\right)
\end{equation}
where $K_h$ is the matrix of keys and $d_{k}$ their dimension.

Attention weights $a_{h,i} = [a_{h,i,0},...,a_{h,i,d_s}]$, where $a_{h,i,j} \in [0,1]$ and $\sum_j{a_{h,i,j}} = 1$, can be seen as a soft-indexing over the values. 
Since the values are projections of the tokens themselves, each weight in self-attention measures the contribution of its token to the attention head and, in turn, to the new token representation.
We provide additional details to the self-attention mechanism in Appendix~\ref{sec:self-attention}.

\paragraph{Attention entropy.}

\begin{figure}[!tp]
    \centering
    \includegraphics[width=.9\linewidth]{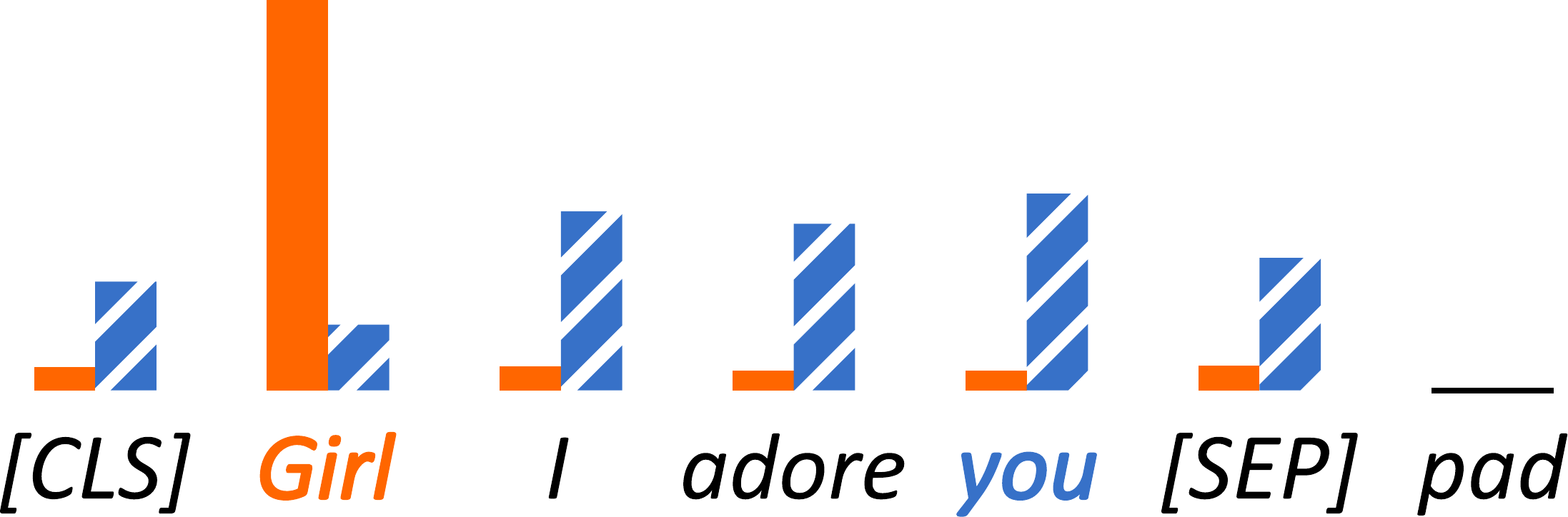}
    \caption{Self-attention distribution on tokens \textit{Girl} (solid orange) and \textit{you} (shaded blue). Attention for \textit{Girl} is concentrated on its representation: its entropy is low. Attention for \textit{you} is spread: its entropy is high.}
    \label{fig:entropy_example}
\end{figure}


Information \textit{entropy} was first introduced in~\newcite{6773024}, and measures the average information content of a random variable $X$ with the set $[x_0,...,x_n]$ of possible outcomes.
It is defined as
\begin{equation}
    H(X) = - \sum_i{P(x_i) \log{P(x_i)}}
\end{equation}

Following \newcite{ghader-monz-2017-attention}, we compute the entropy in the self-attention heads by interpreting each token's attention distribution as a probability mass function of a discrete random variable.
The input embeddings are the possible outcomes, and the attention weights their probability.

For the sake of simplicity, we now discuss the computation of attention entropy of a single token in a standard transformer encoder.
Attention weights are first averaged over heads by defining $a'_{i,j} = \frac{1}{h} \sum_h{a_{h,i,j}}$ as the mean attention that the token at position $i$ pays to the token at position $j$. Then, we define a probability mass function by applying a softmax operator:
\begin{equation}
    a_{i,j} = \frac{e^{a'_{i,j}}}{\sum_j{e^{a'_{i,j}}}}
\end{equation}

We define the attention entropy as follows
\begin{equation}
    H_{i} = - \sum_{j=0}^{d_s}{a_{i,j} \log{a_{i,j}}}
\label{eq:attention_entropy}
\end{equation}

Intuitively, attention entropy measures the degree of contextualization while constructing the model's upper level's embedding. A large entropy suggests that a wider context contributes to the new embedding, while a small entropy tells the opposite: only a few tokens are deemed relevant. 
From a broader viewpoint, contextualized tokens improve the information passage between continuous layers by re-distributing the information content for every unit involved.

Figure~\ref{fig:entropy_example} shows a toy example of self-attention distributions for two arbitrary tokens. Solid orange bars correspond to $a_{\text{Girl},j}$, while shaded blue bars correspond to $a_{\text{you},j}$. The toy example illustrates the correlation between attention distributions and entropy. The representation of \textit{you} uses a wider context and, thus, it has a higher attention entropy.
Note that, if present, we discard padding tokens from the attention entropy computation. Conversely, we include special tokens when required by the downstream task.

\paragraph{EAR in BERT.}

\begin{figure}[!t]
    \centering
    \includegraphics[width= .9 \linewidth]{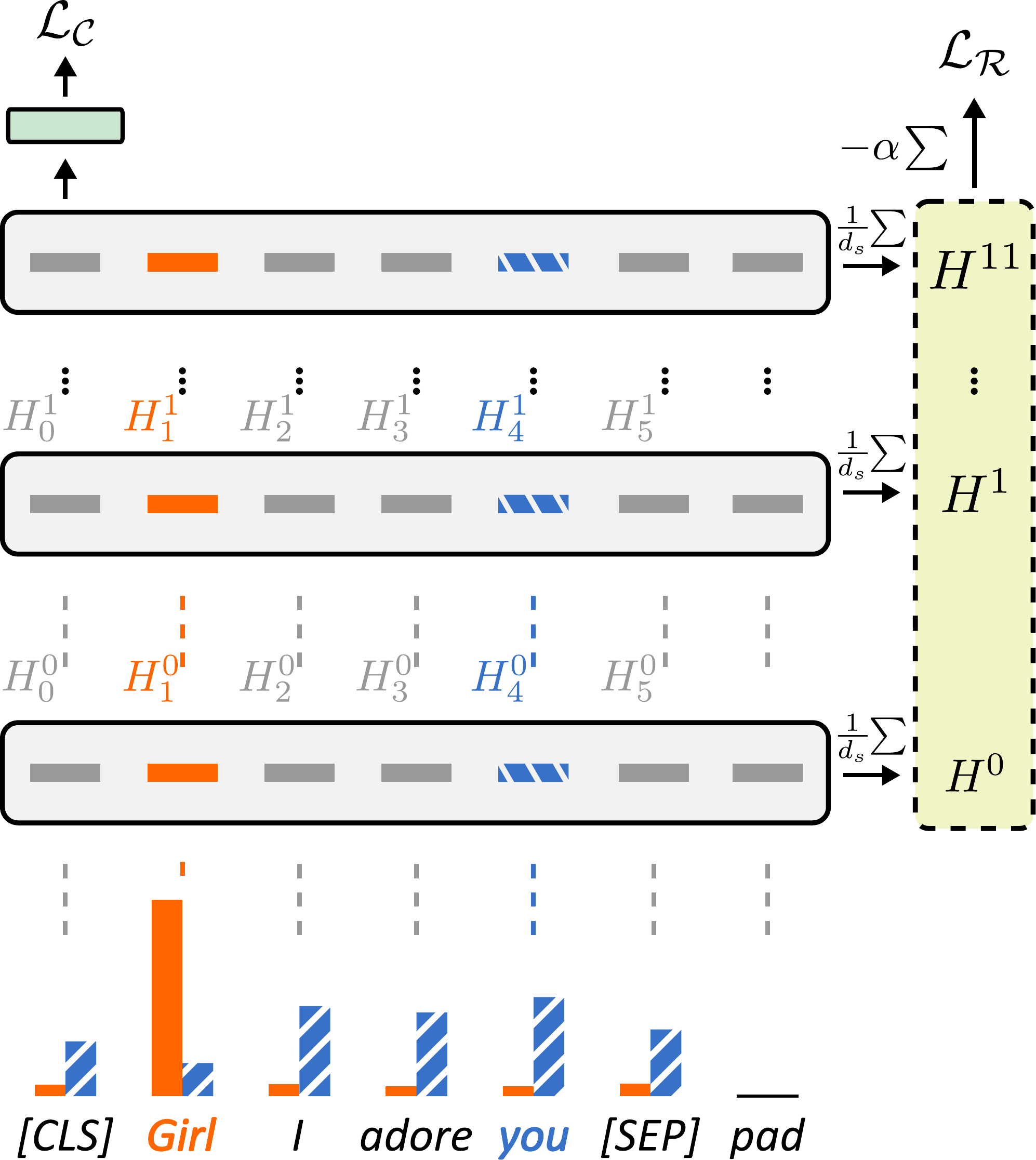}
    \caption{Overview of \placeholder. Grey boxes are Transformer layers. Each builds a token with attention entropy $H_i^{\ell}$. 
    Right green box pools layer-wise contextualization contributions and outputs regularization loss.
    First layer self-attention distribution (bottom) shown for \textit{you} (shaded blue) and \textit{Girl} (solid orange).}
    \label{fig:model_overview}
\end{figure}

We introduced attention entropy as a proxy for the degree of contextualization of token representations above. Following this intuition, we propose BERT with EAR mitigation (\placeholder), a novel model trained to learn tokens with maximal self-attention entropy over the input sequence. We fine-tune \placeholder in the downstream task of hate speech detection. Note, though, that the approach is feasible for any classification task.
In classification models, having more contextualized tokens avoids individual terms driving the classification outcome because they got over-attentioned.

Although \reg is applicable to any Transformer-based model, we base our approach here on the BERT \cite{devlin-etal-2019-bert} base architecture. BERT provides an informative case study, given the number of architectures it has spawned and the recent interest in its attention patterns \cite{clark-etal-2019-bert, kovaleva-etal-2019-revealing, serrano-smith-2019-attention}.
BERT consists of twelve stacked transformer encoders, each running self-attention on the output of the previous encoder. In \placeholder, we build new tokens with the maximal information content coming from the previous layer for every transformer layer in the architecture. Using Equation~\ref{eq:attention_entropy}, we first compute the attention entropy of each token in the input sentence. We then take their mean and define the \textit{average contextualization} for the \textit{$\ell$-th} layer as
\begin{equation}
    H^{\ell} = \frac{1}{d_s} \sum_{i=0}^{d_s}{H_i^{\ell}}
\end{equation}

\noindent
where $H_i^{\ell}$ is the attention entropy of the token at position $i$, and $d_s$ is the length of the input sequence (excluding the padding tokens but including the [CLS] and [SEP] special tokens). Finally, we introduce a new regularization term to the model loss to maximize the entropy at each layer:
\begin{equation}
    \mathcal{L} = \mathcal{L}_C + \mathcal{L}_R ,\quad \mathcal{L}_R = - \alpha \sum_{l}{H^{\ell}}
\label{eq:loss}
\end{equation}
$\mathcal{L}$ is the total loss, $\mathcal{L}_C$ and $\mathcal{L}_R$ are the classification and regularization loss, respectively, and $\alpha \in \mathbb{R}$ is the regularization strength. As in previous work, $\mathcal{L}_C$ is the Cross Entropy loss obtained with a linear layer on top of the last encoder as a classification head. It receives the [CLS] embedding and outputs the probability of the positive class (Hate).

The new regularization term $\mathcal{L}_R$ frames the task of maximal contextualization learning in the network. This framing has several advantages over existing approaches. First, it is a sum of differentiable terms and is hence differentiable. We can thus optimize \placeholder with classical back-propagation updates.
Second, the regularization is agnostic to specific identity terms. It instead induces the network to learn contextualized tokens globally. This induction is crucial to regularize biased terms that might not be known in advance.  
Finally, note that the $\mathcal{L}_R$ pools each layer's entropy-based contributions $H^{\ell}$. Each term $H^{\ell}$ is in turn dependent on the sole attention entropy defined in Equation~\ref{eq:attention_entropy}. This makes the setup a general framework not limited to BERT. $\mathcal{L}_R$ can be used to evaluate and maximize the token contextualization in any attention-based architecture.

Figure \ref{fig:model_overview} shows a graphical overview of \placeholder. Each layer provides a contextualization contributing to the loss independently, where layers with a low average contextualization increase the loss the most. Note also that, similarly to \newcite{he2016deep}, $\mathcal{L}_R$ introduces skip connections between layers and the classification head, so shorter paths for the contextualization information to flow.  


\paragraph{Insights from attention entropy.} On the one hand, we use attention entropy maximization to train \placeholder and test its classification and bias mitigation performance. On the other hand, we can leverage attention entropy to automatically extract the tokens with the lowest contextualization, which are the most likely to induce unintended bias.
When a sentence is fed through a model like BERT, we can inspect the attention distribution of its terms\footnote{For complex terms, we average the attention entropy of their sub-words.}. 


We propose to exploit entropy, and hence contextualization, to gain insights into any attention-based model. Given a corpus and a model we want to inspect, we repeatedly query the model with sentences from the corpus and collect each token's attention entropy. Finally, we take each token's mean to measure the impact it has on bias, where lower is worse. Note that the same term can impact bias differently depending on the sentence.


While our approach works for any attention-based model and data set, we test it on fine-tuned classifiers to extract the biased terms learned on the training data set. We discuss this functionality in Section \ref{biased-terms-extraction}.


\section{Experimental settings}

In this work, we consider the problem of \textit{unintended bias}~\cite{dixon2018}: ``\textit{a model contains unintended bias if it performs better for comments containing some particular identity terms than for comments containing others}''.

\paragraph{Datasets.}
Unintended bias is measured on synthetic test sets, artificially generated by filling manually defined contexts with identity terms (e.g., \textit{I hate all \_\_\_}, \textit{I love all \_\_\_}) . By construction, each identity term appears 50\% of the time in hateful contexts and 50\% in non-hateful ones. 
If a model then classifies the instances related to one identity term differently than the others, it means that the model contains unintended bias towards that term, e.g., if every instance containing the term \textit{women} is labelled hateful, independently of the context. Synthetic test sets simulate new data, so a model that has low performance on this set demonstrates low generalization abilities and incapacity to be used in real-world contexts and applications.

We test \placeholder on hate speech datasets with associated synthetic test sets to measure unintended bias.

\textsc{Misogyny (EN)} \cite{ami-evalita2018} is a state-of-the-art corpus for misogyny detection in English. The related synthetic test set \cite{nozza2019unintended} was created via several manually defined templates and synonyms for ``woman'' as identity terms. 

\textsc{Misogyny (ITA)} \cite{fersini2020ami} is the benchmark corpus for misogyny detection in Italian. The synthetic test set has been generated similarly to the English one. This dataset allows us to study \reg's impact on cross-lingual adaptation.

\textsc{Multilingual and Multi-Aspect Hate Speech} (\textsc{MlMA}) \cite{ousidhoum-etal-2019-multilingual} consists of tweets with various hate speech targets. We choose to work on its English part. We use the synthetic test provided in~\newcite{dixon2018}, generated by slotting a wide range of identity terms into manually defined templates.

\input{dataset_distribution}


Table \ref{tab:distributions} reports statistics of the data sets. Alongside the size of train, test, and validation sets, we report also the percentage of hateful instances to show the class balance. Note that \textsc{MlMA} is highly unbalanced with 88\% of instances associated with the hateful class.
Note that the original \textsc{Multilingual and Multi-Aspect} dataset comes in a multi-label, multiple class setting. Following \newcite{ousidhoum-etal-2021-probing}, we used the \textit{Hostility} dimension of the dataset as target label and created a \textit{Hate} binary from it as follows. We considered single-labeled "Normal" instances to be non-hate/non-toxic and all the other instances to be toxic.

To further characterize our data sets, we explore the aspect of selection bias, reporting the measure $B_{2}$ \cite{ousidhoum-etal-2020-comparative}. The metric ranges from 0 to 1 and evaluates how likely topics of the data set are to contain keywords of the data collection. Values above 0.7 demonstrate high selection bias, implying the need for unbiasing procedures.

We report also the size and number of identity terms used in the synthetic test sets. The percentage of hateful content is perfectly balanced (50\%) since each identity term should appear exactly in the same context as the others to measure the unintended bias. See Appendix~\ref{sec:details_experiments} for the list of identity terms and further preprocessing details.

\subsection{Metrics}

We use the weighted and binary F1-score of the hateful class (\textbf{F1\sub{w}} and \textbf{F1\sub{hate}}) as classification metrics. We consider both due to the class imbalance of test sets (see Table~\ref{tab:distributions}).


We compute the unintended bias metrics from \newcite{dixon2018} and \newcite{borkan2019nuanced}. They are computed from differences in the score distributions between instances mentioning a specific identity-term (\textit{subgroup distribution}) and the rest (\textit{background distribution}). The three per-term AUC-based bias scores are:

1) \textit{AUC\sub{subgroup}} calculates AUC only on the data subset of a given identity term. A low value means the model performs poorly in distinguishing between hateful and non-hateful comments that mention the identity term.

2) \textit{Background Positive Subgroup Negative} (\textit{AUC\sub{bpsn}}) calculates AUC on the hateful background examples and the non-hateful subgroup examples. A low value means that the model confuses non-hateful examples that mention the identity term with hateful examples that do not.

3) \textit{Background Negative Subgroup Positive} (\textit{AUC\sub{bnsp}}) calculates AUC on the non-hateful background examples and the hateful subgroup examples. A low value means that the model confuses hateful examples that mention the identity with non-hateful examples that do not.

We report the averaged metrics across identity terms, i.e., \textbf{AUC\sub{subgroup}}, \textbf{AUC\sub{bpsn}}, and \textbf{AUC\sub{bnsp}}.\footnote{Statistical significance and results from~\newcite{lees2020jigsaw} on these metrics could not be computed due to data unavailability and label distribution assumptions.}

\input{hs_tables}

\subsection{Baselines}
We compare \placeholder against the following existing approaches:
(1)    \textit{BERT} \cite{devlin-etal-2019-bert},
(2)    \textit{BERT+SOC mitigation} \cite{kennedy-etal-2020-contextualizing}, 
    where the authors modify BERT's loss to lower the importance weight of identity terms, computed with the Sampling-and-Occlusion (SOC) algorithm \cite{jin2019towards},
(3)    \newcite{nozza2019unintended}, a single-layer neural network architecture based on the Universal Sentence Encoder (USE) representation \cite{cer-etal-2018-universal},
(4)    \newcite{lees2020jigsaw}, a multilingual BERT model fine-tuned on the training data,
(5)    \newcite{ousidhoum-etal-2021-probing}, a classifier based on TF-IDF and Logistic Regression, and
(6) \newcite{zhang-etal-2020-demographics}, a debiasing training framework based on instance weighting.

The \textit{debiased} version proposed in \newcite{lees2020jigsaw} is obtained by training the model on additional samples from Wikipedia articles (assumed to be non-hateful) to balance the distribution of specific identity terms. \newcite{nozza2019unintended} extracted these additional non-hateful samples from an external Twitter corpus \cite{waseem-hovy-2016-hateful}.

To address the impact of different term lists, we also consider two different versions of BERT+SOC mitigation, one where we test the effect of \textit{missing identity terms} and the other where the identity terms are \textit{translated} for adapting to a new language.






\section{Experimental Results}

\begin{figure*}[!ht]
    \centering
    \includegraphics[width=0.9\linewidth]{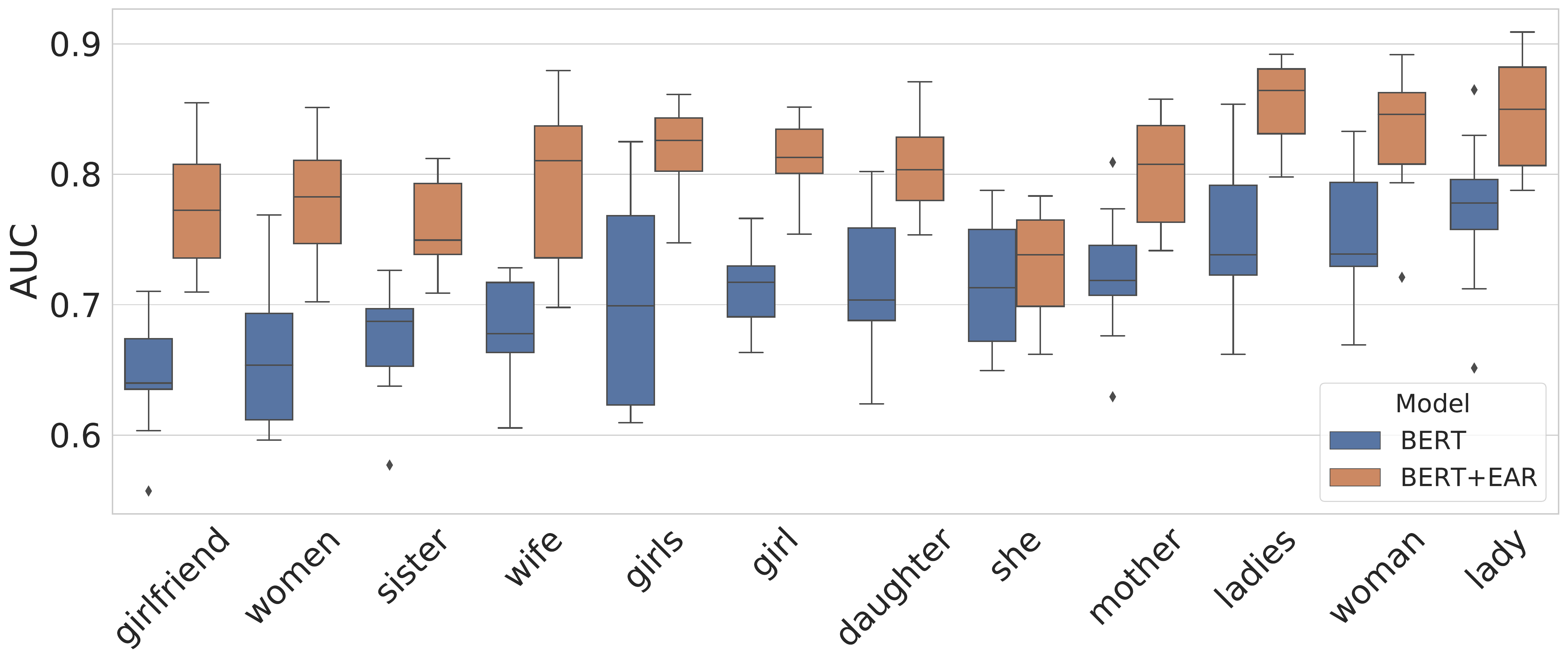}
    \caption{AUC\sub{subgroup} results broken down by identity term on \textsc{Misogyny (EN)}.}
    \label{fig:miso_eng_subgroup_auc}
\end{figure*}


Table \ref{tab:hs_results} shows classification and bias metrics on both synthetic and test set for the three corpora, i.e., \textsc{Misogyny (EN)} (top), \textsc{Misogyny (ITA)} (middle), and \textsc{MlMA} (bottom). 
The top rows in each table section report the performance of hate speech detection models specifically proposed for the respective dataset. The lower rows show the results of baselines and BERT+EAR. 
BERT+SOC mitigation uses the identity terms from \newcite{kennedy-etal-2020-contextualizing} (see Appendix \ref{app:it}), unless a different identity terms lists is specified (e.g., ``BERT+SOC mitigation, translated ITs'').


\placeholder obtains comparable and, in most cases, better performance on all three datasets than all state-of-the-art debiasing approaches, which are based on 
(i)~the knowledge of identity terms and 
(ii)~data augmentation techniques. 
However, identity terms are not always readily available, which severely limits the generalization of those approaches. 
Similarly, there are several drawbacks to data augmentation with (assumed) non-hateful samples containing the identity terms.
1) Data augmentation is expensive. It requires filtering a large dataset (usually Wikipedia) and retraining the model with a much larger set of instances. 
2) Data augmentation with task-specific identity terms requires prior knowledge of those terms, and is therefore limited by the authors' knowledge.
3) The overlap between identity terms in the evaluation set and the augmented data inevitably (but somewhat unfairly) improves the performance on the synthetic dataset.


\placeholder is overall the best debiasing model considering the proposed bias metrics. The only exception is \textsc{Misogyny (EN)}, for which \placeholder has lower \textbf{AUC\sub{bnsp}} and \textbf{AUC\sub{bpsn}} than BERT+SOC mitigation.
The latter's advantage, however, comes with high variability in the results. BERT+SOC mitigation seems more sensitive to random initialization. The 
standard deviation over 10 runs is 37\%, compared to 13\% of \placeholder. 
Figure~\ref{fig:miso_eng_subgroup_auc} shows the AUC\sub{subgroup} metric separately by identity term on \textsc{Misogyny (EN)}. We compare BERT and \placeholder over 10 different initialization runs. \reg improves BERT across all identity terms



Most existing models and AUC-based metrics for unintended bias focus only on the false positives (i.e., hateful instances wrongly recognized as non-hateful). While correctly recognizing hateful instances is important, we believe that the problem of false negatives is equally important. Since \placeholder does not rely on identity term lists, it regularizes terms that impact \textit{both} the positive and negative class. \placeholder obtains an average decrease of 15.04\% in false negative rate compared to BERT and BERT+SOC mitigation. Indeed, the performance difference between BERT+EAR vs.\ BERT and BERT+SOC is mainly due to non-hateful instances ($\sim$95\% of the time). Reducing the impact of overfitting terms like \textit{f*ck} and \textit{p*ssy} in \textsc{Misogyny (EN)} causes BERT+EAR to consider a larger context, and correctly labels them as non-hateful. 

\subsection{Error Analysis}
\label{app:error}

\input{error_analysis}

Table \ref{tab:error_analysis} shows tweets from the \textsc{Misogyny (EN)} data set which have been correctly predicted by BERT+EAR but misclassified by BERT or BERT+SOC. 
These tweets serve as qualitative examples of the effectiveness of forcing the model to attend to a wider context and not overfit to training-specific terms, exploiting the richness of information \cite{nozza-etal-2017-multi}.
The examples are an excerpt of the most common cases where BERT+EAR classifies the non-hateful examples correctly: (1) when slurs or negative words (such as \textit{sk*nk}) are used in a non-hateful context, like slang or lyrics, (2) when many words associated with misogyny appear in the sentence (e.g., \textit{rape}, \textit{abuse}) and (3) when the hateful target is male and the instance should not be classified as misogynous. The use of a wider context by BERT+EAR allows the model identify such non-misogynous instances compared to BERT and BERT+SOC. In particular, BERT+SOC is even more biased in these cases because its debiasing techniques overly rely on specific terms (e.g. \textit{woman}) and increase overfitting to training-specific examples.

\subsection{Impact of predefined identity terms}

We also analyze the impact of predefined identity term lists on performance by evaluating the effect of 
(i)~missing identity terms, and 
(ii)~adapting to a new language where the list is unavailable. 

First, we remove every identity term of BERT+SOC from \textsc{Misogyny (EN)} that appears at least once in the evaluation set, here \textit{women} and \textit{woman} out of 24 terms.
This reflects the real-world case where the identity term list does not contain a specific group present in the data. The significant performance drop resulting from this case (Table~\ref{tab:hs_results}, top, ``missing ITs'') highlights a strong weakness of term-based mitigation strategies.


Second, we analyze the case where identity terms need to be adapted to a new language, e.g., Italian. We translated the English identity terms from BERT+SOC to Italian via Google Translate.\footnote{For gendered Italian words, we kept both the masculine and the feminine (e.g., \textit{muslim} $\rightarrow$ \textit{musulmana}, \textit{musulmano}).} 
Table \ref{tab:hs_results} (middle, ``translated ITs'') shows that the performance is lower than \placeholder. 
A simple translation of predefined identity terms is therefore not an option for cross-lingual settings. This aligns with the findings by \citet{nozza-2021-exposing}, that demonstrated that cross-lingual hate speech detection is limited by the use of non-hateful, language-specific taboo interjections that are not directly translatable.


In sum, we demonstrated that relying on a predefined list of identity terms is a strong limitation for performance and generalizability of the model. In contrast, \placeholder's independence from any predefined terms makes it the ideal model in real-world scenarios.

\section{Extracting overfitting terms}
\label{biased-terms-extraction}

\input{overfitting_terms}

While being the core of \reg, attention entropy serves another purpose. Once standard fine-tuning is concluded (i.e., with no regularization involved), models have overfitted specific terms. We identify  these terms using attention entropy.

To extract the most indicative terms, we replicate training conditions. Specifically, we run inference using all the training data using a fine-tuned checkpoint and a standard BERT tokenizer. We collect attention entropy values for each term and average them over all training instances. Terms with lowest average entropy show the highest overfitting as the model learned them with a narrow context.\footnote{To filter out noise, we report only words with a document frequency higher than 1\%.}

Retrieving these terms after training allows us to gain insights into the domain and language-specific aspects driving the outcome. 


Table \ref{tab:bias_terms} shows the top 10 terms with highest lexical overfitting on the studied datasets extracted from the corresponding fine-tuned model. 
We extract terms strongly correlated with the positive class, e.g., \textit{womens*ck} (97\%), \textit{shut} (96\%), \textit{n*gger} (92\%), \textit{sb*rro} (97\%), \textit{c*lone} (95\%). Note that these terms are \textit{not} frequent in the corpus. Overfitting terms appear with an average document frequency of only 4.7\%, while the most frequent terms have 32.5\% average document frequency across datasets.
These results suggest that the higher the class polarization of a token, the narrower the context BERT will use to learn its representation, and the higher the overfitting.

\section{Related Work}
The first works to study bias measurement and mitigation in neural representation aimed at removing implicit gender bias from word embeddings \cite{bolukbasi2016man,Caliskan183,GargE3635,romanov-etal-2019-whats,ravfogel-etal-2020-null}.
More recently, researchers have started to focus on contextualized sentence representations and effective neural models for understanding the presence and resolution of bias \cite{nozza-etal-2021-honest,ousidhoum-etal-2021-probing}. 

While the majority of proposed approaches focus on data augmentation \cite{dixon2018,nozza2019unintended,sharma2020data,bartl-etal-2020-unmasking,de-vassimon-manela-etal-2021-stereotype}, different approaches have been proposed for bias mitigation intervening directly in the objective function.
\newcite{kennedy-etal-2020-contextualizing} proposed to apply regularization during training to the explanation-based importance of identity terms, obtained with Sampling-and-Occlusion (SOC) explanations \cite{jin2019towards}.
\newcite{kaneko-bollegala-2021-debiasing} proposed a method for debiasing pre-trained contextual representation by retaining the learned semantic information for gender-related words (e.g., \textit{she}, \textit{woman}, \textit{he}, \textit{man}) and simultaneously removing any stereotypical biases in the pre-trained model.
\newcite{zhou-etal-2021-challenges} exploited debiasing methods for natural language understanding \cite{clark-etal-2019-dont} to explicitly determine how much to trust the bias given the input. \newcite{vaidya2020empirical} proposed a multi-task learning model for predicting the presence of identity terms alongside the toxicity of a sentence.

The main drawback of all aforementioned works is their strict reliance on a set of predefined identity terms. This list can be either defined manually by experts or extracted a-priori from the data set. In both cases, the subsequent debiasing models will be strongly affected by these biased terms, limiting the applicability of the trained model to new data. This is a severe limitation, since it is not always possible to retrain a model on new data to reduce bias, resulting in limited use in real-world cases.

\section{Conclusion}

We introduce \reg, a regularization approach applicable to any attention-based model. Our approach does not require any a-priori knowledge of identity terms, e.g., lists. This feature (i) allows us to generalize to different languages and contexts, and (ii) avoids neglecting important terms. Thus, it prevents the introduction of further bias. As part of the training procedure, \reg also discovers the impact of relevant domain-specific terms. 
This automatic term extraction provides researchers with an analysis tool to improve data collection and bias mitigation approaches.

\reg, applied to BERT, reliably classifies data with competitive performance and substantially improves various bias metrics.
\placeholder generalizes better to new domains and languages than similar methods. 


In future work, we will apply EAR-based models to different downstream tasks to both improve bias mitigation and automatically extract biased terms.


\section*{Acknowledgments}
We would like to thank the anonymous reviewers and area chairs for their suggestion to strengthen the paper.
This research is partially supported by funding from the European Research Council (ERC) under the European Union’s Horizon 2020 research and innovation program (No.\ 949944, INTEGRATOR), and by Fondazione Cariplo (grant No. 2020-4288, MONICA).
DN, and DH are members of the MilaNLP group, and of the Data and Marketing Insights Unit at the Bocconi Institute for Data Science and Analysis. EB is member of the DataBase and Data Mining Group (DBDMG) at Politecnico di Torino. GA did part of the work as a member of the DBDMG and is currently a member of MilaNLP. Computing resources were partially provided by the SmartData@PoliTO center on Big Data and Data Science.

\section*{Ethical Considerations}

In this paper, we propose term attention entropy as a proxy for unintended bias in attention-based architectures. Our approach allows us to extract, for a given classifier and data set, a list of terms that induce most of the bias in the model. While this list is intuitive and easy to obtain, we would like to point out some ethical dual-use considerations.

The process of collecting the list is a data-driven approach, i.e., it is strongly dependent on the task, collected corpus, term frequencies, and the chosen model.
Therefore, the list might lack specific terms or include terms that do not strictly perpetrate harm, but are prevalent in the sample.
Because of these twin issues, the resulting lists should \textit{not} be read as complete or absolute. We discourage users from developing new models based solely on the extracted terms. We want, instead, the terms to stand as a starting point for debugging and searching for potential bias issues in the task at hand, be it in data collection or model development. 

Further, while the probability is low, we can not exclude the possibility that future users run \reg on other tasks and data sets to derive private information or profile vulnerable groups.

\bibliography{anthology,custom}
\bibliographystyle{acl_natbib}

\clearpage
\appendix

\input{appendix}

\end{document}

%% file: dataset_distribution.tex
\begin{table}[tb!]
\centering
\small
\begin{tabular}{@{}lccc@{}}
\toprule
                 & \specialcell{\textsc{Misoginy}\\(\textsc{EN})} & \specialcell{\textsc{Misoginy}\\(\textsc{IT})}  &\specialcell{\textsc{MlMA}} \\ \midrule
\# Train                    & 4,000 & 5,000 & 5082 \\
\# Test                     &   1,000 & 1,000 & 565 \\
\% Validation               &   10 & 10 & 10 \\
\% Hate (train, test)                 & 45, 46          & 47, 53      & 88, 88     \\

$B_{2}$ & 0.858 & 0.852 & 0.881 \\  \midrule

\# Synthetic                     &   1,464 & 1,908 & 77,000 \\
\# Identity terms                       & 12            & 18      & 50      \\
\% Hate (Synthetic)  & 50                 & 50          & 50    
\\ \bottomrule
\end{tabular}
\caption{Statistics of the data sets.}
\label{tab:distributions}
\end{table}

%% file: hs_tables.tex
\begin{table*}[htb!]
\adjustbox{max width=\textwidth}{%
\small
\centering
\begin{tabular}{@{}llllll|ll@{}}
\toprule
 & \multicolumn{5}{c|}{\textbf{Unintended bias (synthetic)}} & \multicolumn{2}{c}{\textbf{test}} \\
 & \textbf{AUC\sub{subgroup}} & \textbf{AUC\sub{bnsp}} & \textbf{AUC\sub{bpsn}} & \textbf{F1\sub{w}} & \textbf{F1\sub{hate}} & \textbf{F1\sub{w}} & \textbf{F1\sub{hate}} \\ \midrule
\newcite{nozza2019unintended}, no mitigation & 49.83 & 49.83 & 49.83 & 49.97 & 51.33 & \textbf{72.29} & \textbf{71.62} \\
\newcite{nozza2019unintended}, debiased & 50.27 & 50.21 & 50.21 & 45.40 & 29.31 & 71.43 & 69.37 \\
\midrule
\newcite{zhang-etal-2020-demographics} & 69.99 & 62.19 & 62.19 & 43.01 & 66.70 & 31.35 & 63.21 \\ 
BERT, no mitigation & 70.97 & 66.62 & 66.62 & 58.19 & 64.61 & 69.60 & 70.21 \\
BERT+SOC mitigation & 78.11 & \textbf{76.60} & \textbf{76.60} & 51.88 & 58.89 & 57.39 & 60.47 \\
BERT+SOC mitigation, missing ITs & 68.58 & 67.38 & 67.38 & 38.49 & 41.38 & 51.14 & 43.65 \\
\placeholder & \textbf{80.08} & 75.18 & 75.18 & \textbf{62.59} \sigberthigh \sigbertsochigh & \textbf{70.58} \sigberthigh \sigbertsochigh & 70.90 \sigbertsochigh & 70.83 \sigbertsochigh\\ 
\bottomrule

\midrule

\newcite{lees2020jigsaw}, debiased & - & - & - & \textbf{47.00} & 58.58 & 79.87 & 82.45 \\
\midrule
\newcite{zhang-etal-2020-demographics} & 48.10 & 48.29 & 48.29 & 33.33 & \textbf{66.66} & 33.54 & 66.69  \\
BERT, no mitigation & 47.30 & 47.54 & 47.54 & 39.72 & 61.17 & 81.57 & 83.56 \\
BERT+SOC mitigation, translated ITs & 45.54 & 45.88 & 45.88 & 46.34 & 51.62 & 80.28 & 81.73  \\
\placeholder & \textbf{48.59} & \textbf{48.65} & \textbf{48.65} & 40.64 & 62.71 \sigberthigh \sigbertsochigh & \textbf{83.29} \sigberthigh \sigbertsochigh & \textbf{84.68} \sigbertlow \sigbertsochigh  \\ \bottomrule

\midrule
 
\newcite{ousidhoum-etal-2021-probing}, no mitigation & 63.87 & 60.80 & 61.10 & 33.33 & 66.66 & 82.84 & \textbf{93.80} \\
\midrule
\newcite{zhang-etal-2020-demographics} & 74.14 & 64.74 & 65.76 & 33.33 & 66.66 & 82.84 & 93.79  \\
BERT, no mitigation & 69.38 & 67.12 & 67.12 & \textbf{50.24} & 39.65 & 64.70 & 70.14 \\
BERT+SOC mitigation & 56.15 & 55.83 & 55.58 & 33.79 & 59.89 & 76.49 & 86.24 \\
\placeholder & \textbf{74.31} & \textbf{71.43} & \textbf{71.25} & 40.09 & \textbf{67.45} \sigberthigh \sigbertsochigh 
& \textbf{83.05} \sigberthigh \sigbertsochigh & 91.88 \sigberthigh \sigbertsochigh  \\

\bottomrule
\end{tabular}} 
\caption{Results (in \%) on \textsc{Misogyny (EN)} (top), \textsc{Misogyny (ITA)} (middle), and  \textsc{MlMA}. Significance of \placeholder over BERT without mitigation (\sigberthigh:~$p\leq~0.01$) and BERT with SOC mitigation (\sigbertsochigh: $p\leq 0.01$).}
\label{tab:hs_results}
\end{table*}

%% file: error_analysis.tex
\begin{table*}[ht]
\small
\begin{tabular}{@{}p{8.5cm}llll@{}}
\toprule
text                                                                                                                          & Ground truth & BERT & BERT+SOC & BERT+EAR \\ \midrule
I'm just a sk*nk for understanding the basics of life!                                                                       & 0           & 1    & 1        & 0        \\ \\

You're such a f*cking hoe, I love it - the new Kanye and Lil Pump I kings make women feel comfortable about their sexuality.   & 0           & 1    & 1        & 0    \\ \\

GIRL, YOU’RE   HYSTERICAL. I AM DANCING SO HAPPY FOR   TODAY      & 0           & 0    & 1        & 0        \\ \\
\#metoo I'm a victim of rape, abuse and harrassment. Every woman who had any these experiences. & 0           & 1    & 1        & 0        \\ \\
some people at school drive me insane. like cool b*tch! im depressed too!! doesnt mean im a f*cking c*nt   & 0           & 1    & 1        & 0        \\ \\

@male\_user And you are a hysterical k*nt.                                                                                     & 0           & 0    & 1        & 0        \\ \\

@male\_user F*ck you p*ssy                                                                                  & 0           & 1    & 1        & 0  \\

\bottomrule
\end{tabular}
\caption{Examples of \textsc{Misogyny (EN)} tweets misclassified by BERT or BERT+SOC, and correctly classified by BERT+EAR. Next to the tweet text, we report the ground truth label and the prediction of each model. Exact phrasing changed to protect privacy.}
\label{tab:error_analysis}
\end{table*}

%% file: overfitting_terms.tex
\begin{table*}[!ht]
\small
\centering
\begin{tabular}{@{}ll@{}}
\toprule
\textbf{Dataset}   & \textbf{Overfitting terms} \\      \midrule                                        
\textsc{Misogyny (EN)} & girls, womens*ck, hoes, c*ck, shut, stupid, hoe, p*ssy, trying, f*ck \\
\textsc{Misogyny (ITA)} & pezzo, bel, bellissima, scoperei, p*ttanona, zitta, sb*rro, t*ttona, bella, c*lone \\ 
 & \textit{(piece, nice, very nice, I'd f*ck, sl*t, shut up, c*m, b*sty, beautiful, fat*ss)} \\
\textsc{MlMA} & n*gger, n*gro, shut, chong, ching, d*ke, okay, sp*c, tw*t, f*ggot \\
\bottomrule
\end{tabular}
\caption{Terms with highest lexical overfitting identified using attention entropy.}
\label{tab:bias_terms}
\end{table*}

%% file: appendix.tex
\label{sec:appendix}

\section{Details on self-attention in Transformers}
\label{sec:self-attention}


The Transformer \cite{NIPS2017_3f5ee243} is the building block of many recent neural language models. A Transformer model consists of two connected encoder and a decoder units which align a source and a target sequence. Differentiating from the original formulation, large language models, such as BERT, drop the encoder and use the remaining encoder to process a single input sequence.   

A transformer encoder consists of a multi-head self-attention block and a position-wise, fully connected feed forward neural network. Both the self-attention block and the feed forward network adopt a residual skip connection and batch normalization. We provide details for a standard forward pass in the encoder.
In attention blocks, the multi-head output is computed with Scaled Dot-Product Attention between a set of queries and keys of dimension $d_k$, and a set of values of dimension $d_v$. Let $Q$, $K$ and $V$ be the respective matrix representations. The attention is then computed as
\begin{equation*}
    \operatorname{Attention}(Q, K, V) = \operatorname{softmax}\left(\frac{Q K^{T}}{\sqrt{d_{k}}}\right) V
\label{eq:global_attention}
\end{equation*}

To improve expressiveness, the operation is performed on $N$ different, independent linear projections of the same queries, keys and values, so that $N$ attention heads are produced. The heads are then concatenated, projected back to the original input space, and finally fed through the fully connected neural network to produce the next layer embeddings. Let $E = [{e_0,...,e_{d_s}}]$ be the sequence of input embeddings\footnote{The input embeddings for the first layer are the static token embeddings plus their position encoding.}, with $e_i \in \mathbb{R}^{d_m}$. In the specific case of a transformer encoder, queries, keys and values correspond to the input embeddings - i.e. $Q = K = V = E$. As such, the output of the multi-head self-attention block is computed applying the previously presented Equation to the $N$ token projections, concatenating and projecting back to the original space:

\begin{equation*}
\operatorname{MultiHead}(Q, K, V) = \left(\text{o}_{0} || \ldots ||  \text{o}_{N}\right) W^{O}    
\end{equation*}

\noindent
where
\begin{equation*}
\text {o}_{h} = \text{Attention}\left(Q W_{h}^{Q}, K W_{h}^{K}, V W_{h}^{V}\right)
\end{equation*}

\noindent
and $W^{O}$ and each $W^{Q}_h$, $W^{K}_h$, $W^{V}_h$ are projection matrices.

\section{Experimental setup}
\label{sec:details_experiments}

\paragraph{Hyper-parameters}
All our experiments use the Hugging Face transformers library~\cite{wolf-etal-2020-transformers}. We base our models and tokenizers on the \texttt{bert-base-uncased} checkpoint for English tasks and on the \texttt{dbmdz/bert-base-italian-uncased} checkpoint for Italian.
We pre-process and tokenize our data using the standard pre-trained BERT tokenizer, with a maximum sequence length of 120 and right padding. 
We train all models with the following hyperparameters: batch size=64, learning rate=$0.00002$, weight decay=0.01, learning rate warmup steps=10\%, full precision, maximum number of training epochs=30, and early stopping on non-improving validation loss after 5 epochs. Table \ref{tab:hs_results} report results of \placeholder trained for 20 epochs with no early stopping, and regularization strength $\alpha=0.01$. We chose the latter parameters with grid search on $\alpha \in [0.0001, 0.001, 0.01, 0.1, 1]$ and $\text{epochs} \in [10, 20, 30, 40, 50]$.
When fine-tuning on \textsc{Multilingual and Multi-Aspect}, we use a weighted cross-entropy classification loss ($\mathcal{L}_C$) to discount class unbalance. Specifically, we normalize the loss for data points belonging to class $C$ by the prior probability of $C$, evaluated as its relative frequency in the training set. 

For \newcite{kennedy-etal-2020-contextualizing}, \newcite{nozza2019unintended}, \newcite{lees2020jigsaw}, and \newcite{ousidhoum-etal-2021-probing}, we kept all the parameters as specified by the respective authors. Please refer to our repository (\repository) for further details or the respective publications.

We trained all models with 10 different initialization seeds per parameter configuration and averaged over them to obtain stable results and meaningfully compute significance. 

\paragraph{Statistical significance}
We compute the statistical significance of \placeholder over BERT and BERT with SOC mitigation via bootstrap sampling, following \newcite{sogaard-etal-2014-whats}, using \sigbertlow~and \sigbertsoclow~(and their filled counterparts for a stronger significance) symbols, respectively. We use 1000 bootstrap samples and a sample size of \%20. For Hate Speech, significance can only be computed on F1-scores, since bias metrics require an assumption about the label distribution across identity terms that is not given.

\paragraph{Selection bias}
We computed the \textbf{B\sub{2}} metric following \newcite{ousidhoum-etal-2020-comparative}. Specifically, we run the authors' code on each of our training dataset, using the query keywords used to sample each dataset. In case of queries composed of multiple words, we split and considered them separate keywords.

\paragraph{Dataset preprocessing}
The original \textsc{Multilingual and Multi-Aspect} dataset comes in a multi-label, multiple class setting. Following \newcite{ousidhoum-etal-2021-probing}, we used the \textit{Hostility} dimension of the dataset as target label and created a \textit{Hate} binary from it as follows. We considered single-labeled "Normal" instances to be non-hate/non-toxic and all the other instances to be toxic.

\paragraph{Computation time}
We report NVIDIA Tesla V100 PCIE-16GB -equivalent computation time for the tested models. Averaging across the three presented data sets, training and evaluating 10 seeds of \placeholder (without early stop) requires 22 hours, compared to 72 hours for BERT+SOC and 7 hours for BERT. The regularization of attention entropy does not affect the computation time by a significant amount.

\paragraph{CO\textsubscript{2} emission}

Experiments were conducted using a private infrastructure, which has an estimated carbon efficiency of 0.432 kgCO$_2$eq/kWh. A cumulative of 319 hours of computation was performed on the hardware of type Tesla V100-PCIE-16GB (TDP of 300W).
Total emissions are estimated to be 41.34 kgCO$_2$eq.

Estimations were conducted using the \href{https://mlco2.github.io/impact#compute}{Machine Learning Impact calculator} presented in \cite{lacoste2019quantifying}.

\section{List of identity terms}
\label{app:it}

In the following, we report the list of identity terms used in the considered data sets and methods.

\begin{quote}

\cite{kennedy-etal-2020-contextualizing}: \textit{\textit{muslim}, \textit{jew}, \textit{jews}, \textit{white}, \textit{islam}, \textit{blacks}, \textit{muslims}, \textit{women}, \textit{whites}, \textit{gay}, \textit{black}, \textit{democrat}, \textit{islamic}, \textit{allah}, \textit{jewish}, \textit{lesbian}, \textit{transgender}, \textit{race}, \textit{brown}, \textit{woman}, \textit{mexican}, \textit{religion}, \textit{homosexual}, \textit{homosexuality}, \textit{africans}}

\cite{nozza2019unintended}: \textit{woman}, \textit{women}, \textit{daughter}, \textit{girl}, \textit{girls}, \textit{mother}, \textit{she}, \textit{wife}, \textit{lady}, \textit{ladies}, \textit{girlfriend}, \textit{sister}


\cite{fersini2020ami}: \textit{nonne}, \textit{matrone}, \textit{mamme}, \textit{casalinghe}, \textit{compagne}, \textit{morose}, \textit{femmine}, \textit{donne}, \textit{fidanzate}, \textit{nonna}, \textit{matrona}, \textit{casalinga}, \textit{morosa}, \textit{femmina}, \textit{mamma}, \textit{donna}, \textit{fidanzata}, \textit{compagna}

\cite{dixon2018}: \textit{lesbian}, \textit{gay}, \textit{bisexual}, \textit{transgender}, \textit{trans}, \textit{queer}, \textit{lgbt}, \textit{lgbtq}, \textit{homosexual}, \textit{straight}, \textit{heterosexual}, \textit{male}, \textit{female}, \textit{nonbinary}, \textit{african}, \textit{african american}, \textit{black}, \textit{white}, \textit{european}, \textit{hispanic}, \textit{latino}, \textit{latina}, \textit{latinx}, \textit{mexican}, \textit{canadian}, \textit{american}, \textit{asian}, \textit{indian}, \textit{middle eastern}, \textit{chinese}, \textit{japanese}, \textit{christian}, \textit{muslim}, \textit{jewish}, \textit{buddhist}, \textit{catholic}, \textit{protestant}, \textit{sikh}, \textit{taoist}, \textit{old}, \textit{older}, \textit{young}, \textit{younger}, \textit{teenage}, \textit{millenial}, \textit{middle aged}, \textit{elderly}, \textit{blind}, \textit{deaf}, \textit{paralyzed}

\end{quote}